\documentclass[letterpaper]{article} % DO NOT CHANGE THIS
\usepackage{aaai23}
\usepackage{times}  % DO NOT CHANGE THIS
\usepackage{helvet}  % DO NOT CHANGE THIS
\usepackage{courier}  % DO NOT CHANGE THIS
\usepackage[hyphens]{url}  % DO NOT CHANGE THIS
\usepackage{graphicx} % DO NOT CHANGE THIS
\usepackage{natbib}  % DO NOT CHANGE THIS AND DO NOT ADD ANY OPTIONS TO IT
\usepackage{caption} % DO NOT CHANGE THIS AND DO NOT ADD ANY OPTIONS TO IT
\usepackage{algorithm}
\usepackage{algorithmic}
\usepackage{appendix}
\usepackage{newfloat}
\usepackage{listings}
\DeclareCaptionStyle{ruled}{labelfont=normalfont,labelsep=colon,strut=off} % DO NOT CHANGE THIS
\floatstyle{ruled}
\newfloat{listing}{tb}{lst}{}
\floatname{listing}{Listing}
\usepackage{subfig}
\usepackage{booktabs}
\usepackage{multirow}
\usepackage{amsmath,amssymb}
\usepackage{xspace}
\usepackage{hyperref}

\usepackage{amsmath,amsfonts,bm}

\def\eqref#1{equation~\ref{#1}}
\def\1{\bm{1}}

\def\vf{{\bm{f}}}

\def\vp{{\bm{p}}}

\def\mC{{\bm{C}}}

\def\mF{{\bm{F}}}

\def\mN{{\bm{N}}}

\def\mP{{\bm{P}}}

\def\mW{{\bm{W}}}

\DeclareMathAlphabet{\mathsfit}{\encodingdefault}{\sfdefault}{m}{sl}
\SetMathAlphabet{\mathsfit}{bold}{\encodingdefault}{\sfdefault}{bx}{n}
\newcommand{\tens}[1]{\bm{\mathsfit{#1}}}

\def\tE{{\tens{E}}}
\def\tF{{\tens{F}}}

\def\tO{{\tens{O}}}

\def\tW{{\tens{W}}}

\makeatletter
\DeclareRobustCommand\onedot{\futurelet\@let@token\@onedot}
\def\@onedot{\ifx\@let@token.\else.\null\fi\xspace}

\def\ie{\emph{i.e}\onedot}

\makeatother

\title{SEFormer: Structure Embedding Transformer for 3D Object Detection}
\author{
    Xiaoyu Feng \textsuperscript{\rm 1},
    Heming Du \textsuperscript{\rm 2},
    Yueqi Duan \textsuperscript{\rm 1},
    Yongpan Liu \textsuperscript{\rm 1},
    Hehe Fan \textsuperscript{\rm 3}
}
\affiliations{
    \textsuperscript{\rm 1} Tsinghua University\\
    \textsuperscript{\rm 2} Australian National University\\
    \textsuperscript{\rm 3} National University of Singapore\\
    feng-xy18@mails.tsinghua.edu.cn
}

\usepackage{bibentry}
\begin{document}

\maketitle

\begin{abstract}
    % Effectively  preserving and encoding structure features from objects in irregular and sparse LiDAR points is a key challenge to 3D object detection on point cloud.
    % Pure CNN based solutions learn fixed spatial kernel weights to encode local points with different directions and distances from the kernel center.
    % However, the fixed and rigid kernel size of convolution makes convolution inevitable to include unrelated points. 
    % The self-attention mechanism in Transformer is able to adaptively search related or similar points and thus preserve the local spatial structure.
    % But Transformer only performs a simply sum on the point features, based on the self-attention, and lacks the direction-distance-oriented structure encoding ability as convolution.
    % To address the above mentioned problem in vanilla Transformer, we propose a Structure-Embedding transFormer (SEFormer), which combines the merits of convolution for structure encoding and Transformer for local structure preservation. 
    % Compared to the self-attention mechanism in traditional Transformer, SEFormer learns different feature transformations for \textit{value} based on the relative directions and distances to the query point.  
    % Then we propose a SEFormer based network for high-performance 3D object detection.
    % Extensive experiments show that the proposed architecture can achieve the state-of-the-art results on Waymo Open Dataset, the largest 3D detection benchmark for autonomous driving. Specifically, SEFormer achieves $78.85\%$ mAP, which is $1.53\%$ higher than existing works.
    Effectively  preserving and encoding structure features from objects in irregular and sparse LiDAR points is a key challenge to 3D object detection on point cloud.
    Recently, Transformer has demonstrated promising performance on many 2D and even 3D vision tasks.
    Compared with the fixed and rigid convolution kernels, the self-attention mechanism in Transformer can adaptively exclude the unrelated or noisy points and thus suitable for preserving the local spatial structure in irregular LiDAR point cloud.
    However, Transformer only performs a simple sum on the point features, based on the self-attention mechanism, and all the points share the same transformation for \textit{value}.  
    Such isotropic operation lacks the ability to capture the direction-distance-oriented local structure which is important for 3D object detection.
    In this work, we propose a Structure-Embedding transFormer (SEFormer), which can not only preserve local structure as traditional Transformer but also have the ability to encode the local structure. 
    Compared to the self-attention mechanism in traditional Transformer, SEFormer learns different feature transformations for \textit{value} points based on the relative directions and distances to the query point.  
    Then we propose a SEFormer based network for high-performance 3D object detection.
    Extensive experiments show that the proposed architecture can achieve SOTA results on Waymo Open Dataset, the largest 3D detection benchmark for autonomous driving. Specifically, SEFormer achieves $79.02\%$ mAP, which is $1.2\%$ higher than existing works. We will release the codes.
\end{abstract}

\section{Introduction}

Point cloud based 3D object detection has attracted more and more attention with the rapid development of autonomous driving and robotics. 
Due to the lack of texture and color information in point cloud, 3D object detection highly depends on the structure information of local areas. 
However, unlike the grid-arranged 2D images, the sparse and irregular nature of LiDAR point clouds makes the local structure often incomplete and noisy. 
Hence, how to effectively extract the required structure feature is still an unsolved problem. 

\begin{figure}[t]
  \centering
  \includegraphics[width=0.45\textwidth]{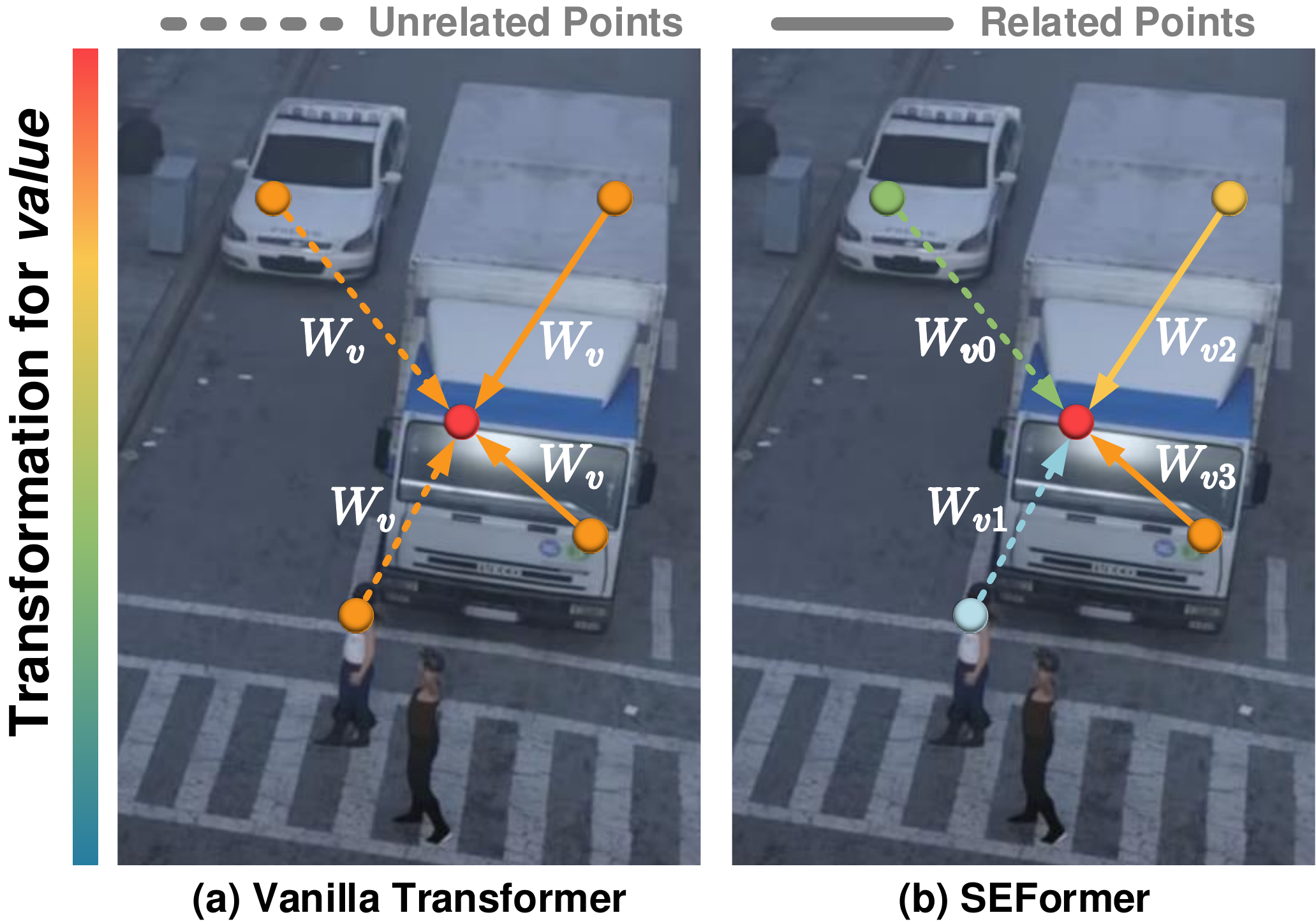}
  \caption{
  Motivation illustration: (a) The self-attention mechanism in Transformer can adaptively exclude the unrelated or noisy points. However, vanilla Transformer shares the same feature transformation for all \textit{value} points. Such isotropic operation ignores the local structure information between the \textit{key} points and the \textit{query} point.
   (b) While in SEFormer, we propose to learn different feature transformations for \textit{value} points based on their relative directions and distances to the \textit{query} point. Hence the local structure can be encoded in the Transformer output. For example, SEFormer can differentiate the front and tail points from a car.
  \vspace{-15pt}
  }
  \label{fig:philosophy}
\end{figure}

Inspired by the success of 2D object detection \citep{ren2015faster, wu2021vector, wu2021instance, shrivastava2016training, he2017mask,sun2021sparse,chi2020relationnet++,dong2022construct,han2022meta,yan2022unsupervised,zhang2022lgd}, convolution rapidly becomes the mainstream operator in 3D object detection.
Traditional convolution based 3D object detection can be divided into two main trends: point \citep{qi2017pointnet, qi2017pointnet++, shi2019pointrcnn, qi2018frustum, yang20203dssd,chen2022sasa,he2022svga,sun2022correlation} and voxel-based solutions \citep{yan2018second, shi2020pv, deng2020voxel, zheng2021se, lang2019pointpillars, shi2022pv, li2021voxel, li2021anchor,song2022jpv,xu2022behind}. 

However, convolution is designed with fixed and rigid kernel sizes and treats all neighboring points equally.
Therefore, it inevitably contains unrelated or noisy points from other objects or backgrounds. 
% Convolution learns a set of different kernels to embed point features from different distances and directions (Fig. \ref{fig:philosophy} (b)).
% It allows convolution to effectively \textbf{encode} the spatial structure of the neighboring local area to the center point.
% Consequently, convolution-based methods have achieved great progress on several benchmarks such as KITTI \citep{geiger2013vision} and Waymo \citep{sun2020scalability}.
% However, convolution is designed with fixed and rigid kernel sizes and treats  all neighboring points equally with the aggregation coefficient ($\alpha = 1$).
% Therefore, it inevitably contains unrelated or noisy points from other objects and  has the limited ability to \textbf{preserve} local structure.
Recently, Transformer \citep{vaswani2017attention} has shown its effectiveness in 3D  vision tasks such as classification, segmentation and object detection \citep{zheng2022hyperdet3d,zhao2021point,zhao2022codedvtr,wang2021anchor,zhen2022deeply,cao2022cf,chen2022lctr}.
Compared with convolution, the self-attention mechanism in Transformer is able to adaptively exclude noisy or irrelevant points from other objects.
We refer to such ability of Transformer as \textbf{structure preserving}.
However, vanilla Transformer shares the same feature transformation for points \textit{value}. 
Such isotropic operation ignores the local structure information in the spatial relationships, e.g. directions and distances, from the center point to its neighbors.
As illustrated in Fig. \ref{fig:philosophy} (a), the points share same transformation.
If we swap the positions of points, the Transformer output still remains the same.
It brings challenge for recognizing the object direction, which is important to 3D object detection.

In this work, we are motivated by the fact that convolution learns a set of different kernels to embed point features from different distances and directions.  
Hence, we design a novel Structure-Embedding transFormer (SEFormer),  which can encode the direction-distance-oriented local into its output.
Compared with vanilla Transformer, the proposed SEFormer learns different transformations for \textit{value} points from different directions and distances. 
Hence the change of local spatial structure can be encoded in the output features. 
We refer to SEFormer's such ability as \textbf{structure encoding}. 
As shown in Fig. \ref{fig:philosophy}(b), the points are embed with different transformations (differently colored arrows). 
Once the points are swapped, the correspondence between the points and transformations is changed. 
Hence, the position change can be encoded in the SEFormer output and provide a clue to accurately recognize the object directions.
As a Transformer, SEFormer can also adaptively \textbf{preserve} the local structure. 
Coupled with the additional structure \textbf{encoding} ability, SEFormer can extract better local structure information. 

Based on the proposed SEFormer unit, we propose a multi-scale SEFormer network to extract local structure description for 3D object detection.
Specifically, we extract \textit{point-} and \textit{object-level} structure features.
Based on the multi-scale features extracted by stacked convolution layers, 
multiple parallel SEFormer blocks, each of which contains stacked SEFormer units with different neighbor search radii, are utilized to extract richer structure features around each sampled embedding point.
Each embedding point provides a point-level structure description of its surrounding local area. 
Then theses point-level embedding features are sent to a SEFormer-based detection head.
The network first predicts multiple potential region proposals. 
Based on stacked SEFormer layers, each proposal integrates its nearby point embedding and outputs a object-level feature embedding.
The final bounding boxes are generated based on such object-level embedding.
Our main contributions are threefold: 
\begin{itemize}
\vspace{-2pt}
    \item We propose SEFormer, a new Structure-Embedding transFormer to capture the local point structure. 
    SEFormer can not only preserve the local structure as traditional Transformer but also have additional ability to encode the local direction-distance-oriented structure. 
    \vspace{-2pt}
    \item Based on the proposed SEFormer unit, we design a new multi-scale 3D object detection framework. 
    With multiple SEFormer blocks, we extract point- and object-level structure features for more accurate detection.
    \vspace{-2pt}
    \item Extensive experiments prove the advantages of our SEFormer. 
    On Waymo Open dataset, we achieve $79.02\%$ mAP, which is $1.2\%$ higher than existing works.
\end{itemize}

\section{Related Work}

\noindent\textbf{3D Object Detection on Point Cloud.} 3D object detection methods on point cloud recently made a giant leap with the advancement of deep neural networks.
According to different representations of point cloud inputs, recent research can be categorized into two main trends: point and voxel-based.

% \noindent\textbf{Multi-view-based Object Detection.}
% For the former trend, \citet{chen2017multi} first generates 3D object proposals from a bird's eye view and then combines these proposals with other two views, \ie, a front view from LiDAR point cloud and an RGB view from images. 
% % Based on MV3D, AVOD~\citep{ku2018joint} improves performance by fusing features from both RGB images and point cloud from bird's eye views through Feature Pyramid Network (FPN). 
% Based on MV3D, AVOD~\citep{ku2018joint} improves feature fusion through Feature Pyramid Network (FPN).
% Afterwards, \citet{yoo20203d} design 3D-CVF to combine transferred features from RGB images and LiDAR point cloud in each region by a cross-view spatial feature fusion strategy. 

\begin{figure*}
\centering
\includegraphics[width=1.0\textwidth]{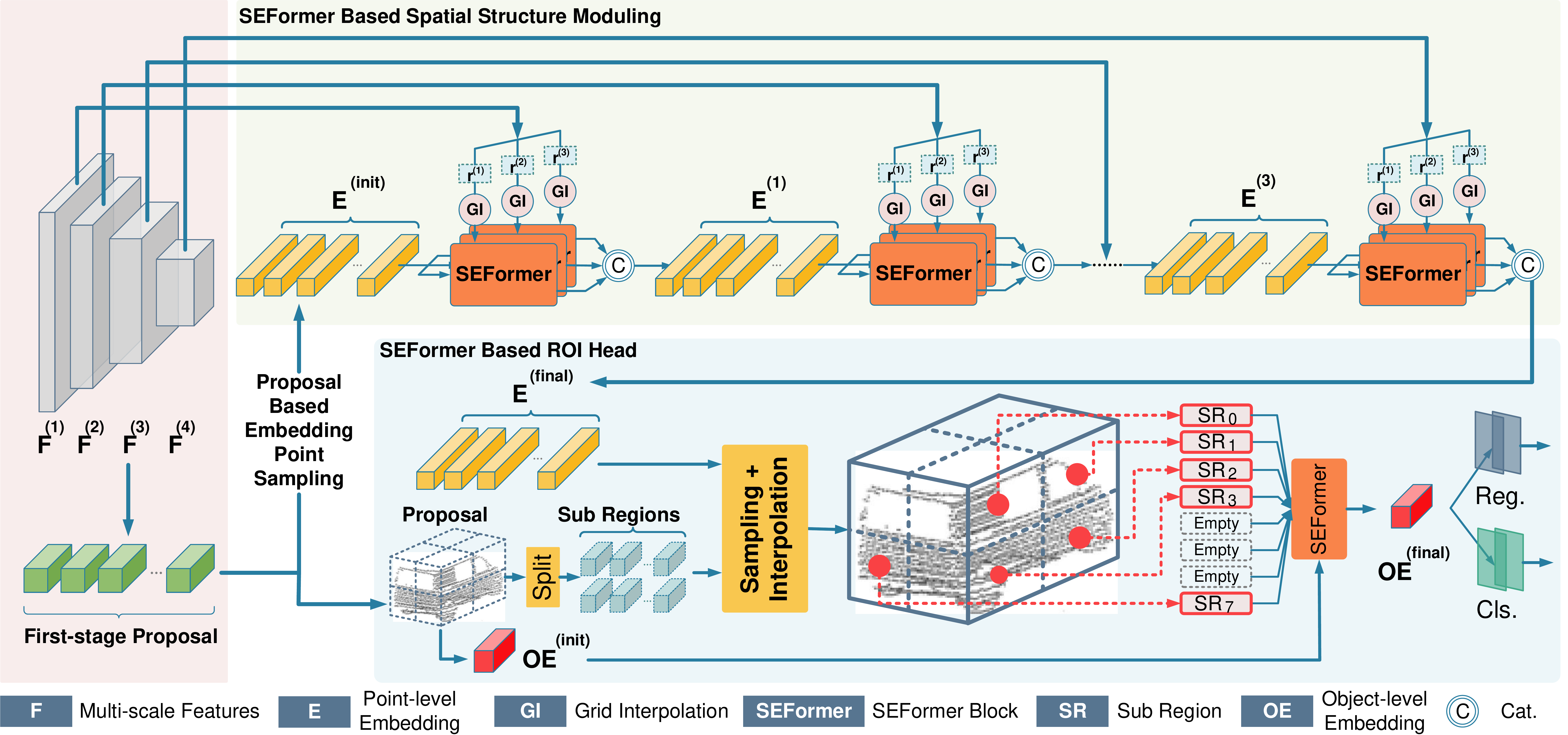}
\caption{An overview of the proposed multi-scale SEFormer network. Stacked convolution layers are first used to extract mutli-scale voxel features. Then a SEFormer based spatial structure moduling aggregates the multi-scale features into several point-level embedding features (yellow cube). A SEFormer based detection head further integrates the point-level embedding with predicted region proposals to generate the object-level embedding features (red cube) for final bounding boxes prediction. }
\label{fig:architecture}
\end{figure*}

(1) Point-based Object Detection. Some works~\citep{qi2019deep,xie2021venet,chen2022sasa} propose to directly process raw point cloud data by adopting point-based backbones, such as PointNet~\citep{qi2017pointnet} and PointNet++~\citep{qi2017pointnet++}. 
% are two of the most popular backbones for point-based object detection methods.
To process a mass of LiDAR points in outdoor environments, \ie, KITTI~\cite{geiger2013vision} and Waymo~\cite{sun2020scalability}, previous point-based approaches~\citep{qi2018frustum,shi2019pointrcnn,yang2019std} usually reduce the number of input points. 
However, such sampling inevitably causes information loss and limits the detection precision.

(2) Voxel-based Object Detection. Voxel-based works \citep{yan2018second,shi2020pv,deng2020voxel,song2022jpv,xu2022behind} transform raw point cloud into compact voxel-grid representation and utilize efficient 3D sparse convolution operator so high-resolution voxelization can be adopted during computation.
In this way, the voxel-based methods often achieve higher performance than the point-based ones. 
In this work, we mainly refer to voxel-based detection when talking about convolution-based works.

\noindent\textbf{Transformer for 3D Object Detection.} 
Many researchers are motivated by recent success of Transformer in computer vision and have tried to introduce Transformer into 3D object detection. 
Point Transformer \citep{zhao2021point} is the pioneering work to apply Transformer in point cloud.
Pointformer \citep{pan20213d} follows their paradigm and designs three Transformers to extract features from different scales. 
While Voxel Transformer \citep{mao2021voxel} combines Transformer with voxel representation and achieves much higher precision.
\citet{fan2022embracing} propose a single-stride Transformer and improves the detection precision on small objects such as pedestrians and cyclists.
However, the vanilla Transformer adopted in such works lacks the ability of local structure encoding.  
Hence we propose SEFormer, a new Transformer, to better extract local structure.  
% Our experiment have shown SEFormer have significant outperformed existing 3D object detection methods on the challenging Waymo benchmark.  

\section{Method}
\begin{figure}[t]
  \centering
  \begin{subfloat}[Attention of Vanilla Transformer]{\includegraphics[width=0.47\textwidth]{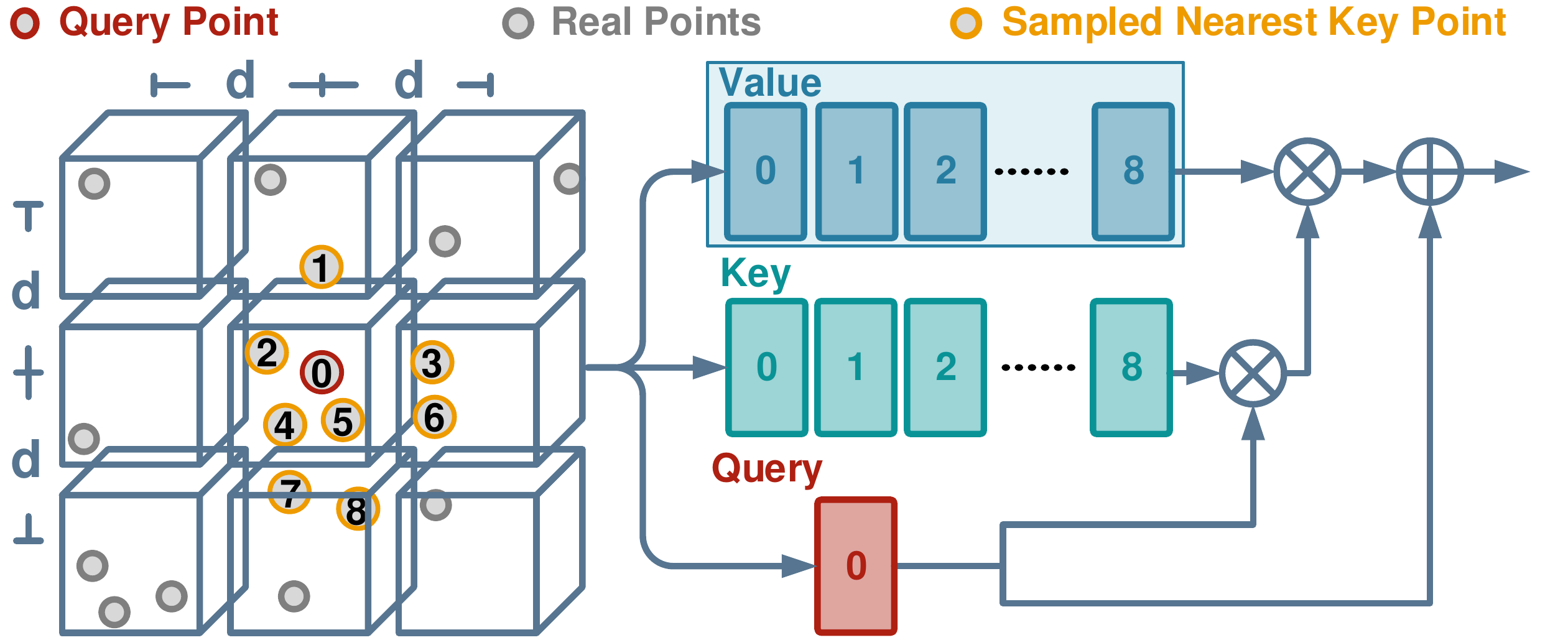}}
  \end{subfloat}
  \vfill
  \begin{subfloat}[Attention of SEFormer]{\includegraphics[width=0.47\textwidth]{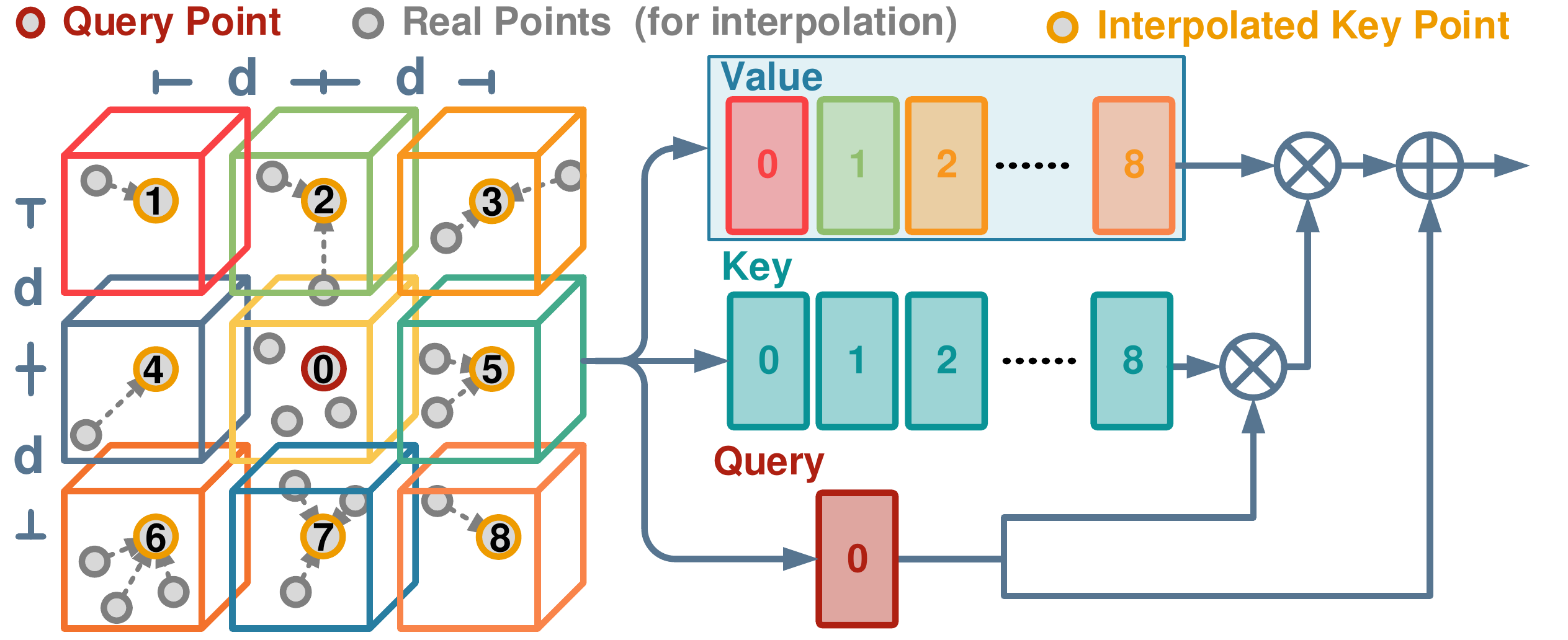}}
  \end{subfloat}
  
  \caption{Architecture comparison of Transformer and SEFormer. 
  (a) In vanilla Transformer, 9 nearest points (8 key points + 1 center query point) are sampled and the points share the same transformation for \textit{value}.
  (b) While in SEFormer, the key points are generated with gird interpolation and different points have their own transformation for \textit{value}. 
%   (a) Vanilla Transformer uses sampling methods like KNN to choose 9 nearest points from the neighboring square area (numbered orange fringed grey points) as key points. 
%   KNN greedily samples the nearest points and there is an obvious bias focusing on the bottom right points. 
%   During calculation, all the sampled points share similar feature transformation for \textit{query}, \textit{key} and \textit{value}. 
%   (b) While SEFormer first adopts grid interpolation to generate 9 anchor points (numbered orange points) as the key points. 
%   Each anchor point is interpolated with its neighboring grey points.
%   Compared to KNN, grid interpolation can forcibly extract points from different directions and provides a better description for local structure.
%   In SEFormer, each key point is transformed independently for \textit{value} (differently colored \textit{value}) based on its direction and distance to the center query point. 
}
  \label{fig:attention}
\end{figure}

In this section, we will first introduce the proposed SEFormer unit, including its motivation and corresponding architecture. 
Then we will present the proposed detection architecture in following sections. 
\vspace{-8pt}
\subsection{Structure Embedding Transformer (SEFormer)}
\vspace{-3pt}

\textbf{Structure Preserving $\&$ Encoding}
\label{sec:motivation}
Before introducing SEFormer, we will first state our main motivation, how to achieve simultaneous structure preserving and encoding. 
Such motivation comes from one key insight we have on two existing operators, convolution and Transformer.

Convolution is the most famous operator in computer vision tasks. 
Because the locality and spatial invariance of convolution well adapt to the inductive bias in images. 
While we propose another important advantage of convolution is that it can \textbf{encode} the structural information of data. 
To illustrate such a point, we first formulate convolution as follows:
\begin{equation}
\label{eq1}
    \vf_{\vp}^{'} = \sum_{\delta} w_{\delta(\vp)} \cdot \vf_{\vp+\delta(\vp)}
    \vspace{-3pt}
\end{equation}
Here $\vf,\vf^{'}$ represents the input and output feature of a convolution layer at center position $\vp$. 
While $\delta$ denotes the relative position between the neighboring points and the center point. 
We decompose convolution as a two-step operator, transformation, and aggregation. 
During transformation, each point will be multiplied by its corresponding kernel $w_{\delta}$. 
Then these points will be simply summed with a fixed aggregation coefficient $\alpha = 1$. 
In convolution, the kernels are differently learned based on their directions and distances to the kernel center. 
Hence convolution can \textbf{encode} the local spatial structure into the output. 
However, in convolution, all neighboring points are equally ($\alpha = 1$) treated during aggregation. 
The mainstream convolution operator adopts a static and rigid kernel but the LiDAR point cloud is often irregular and even incomplete. 
Hence convolution inevitably includes irrelevant or noisy points into the output feature.

Compared with convolution, the self-attention mechanism in Transformer provides a more effective method to \textbf{preserve} the irregular objects' shapes and boundaries in Point Cloud. 
For a point cloud with $N$ points, $\vp = [p_1,\ldots,p_N]$, Transformer computes the response of each point as:
\begin{equation}
\label{eq2}
    \vf_{p}^{'} = \sum_{\delta} \alpha_{\delta(\vp)} (\mW^v \vf_{\vp+\delta(\vp)})
    \vspace{-3pt}
\end{equation}
Here $\alpha_{\delta}$ represents the self-attention coefficients among points in the neighboring area while $\mW^v$ means the value transformation.
We can still decompose Eq. \ref{eq2} into a transformation process with a transformation matrix $\mW^v$ and an aggregation process with attention coefficients $\alpha_{\delta}$. 
The coefficient $\alpha_{ij}$ between point $\vp_i$ and $\vp_j$ can be calculated as $\alpha_{ij} = \frac{exp(e_{ij})}{\sum_{k=1}^N exp(e_{ik})}$.
% \begin{equation}
% \label{eq3}
% \alpha_{ij} = \frac{exp(e_{ij})}{\sum_{k=1}^N exp(e_{ik})}
% \end{equation}
Here $e_{ij} = \frac{(\mW^q \vf_{i})(\mW^k \vf_{j})^{T}}{\sqrt{c}}$ is the scaled dot-produce attention between $\vp_i$ and $\vp_j$ and $\mW^q, \mW^k$ represent the transformation matrix for \textit{query} and \textit{key}.
% \begin{equation}
% \label{eq4}
%     e_{ij} = \frac{(\mW^q \vf_{i})(\mW^k \vf_{j})^{T}}{\sqrt{c}}
% \end{equation}
Compared with the static $\alpha = 1$ in convolution, the self-attention coefficients allow Transformer to adaptively choose the points for aggregation and exclude the influence of unrelated points. 
We call Transformer's such ability as \textbf{structure preserving}. 
However, according to Eq. \ref{eq2}, a same transformation for \textit{value} is shared among all the points in Transformer. 
This means the \textbf{structure encoding} ability, which convolution has, is missed by vanilla Transformer. 

Given the above discussion, we can find that convolution has the ability to \textbf{encode} data structure while Transformer can well \textbf{preserve} the structure. 
Hence a straightforward idea is to design a new operator having both the advantages of convolution and Transformer. 
Hence we propose a Transformer, SEFormer, which can be formulated as:
\begin{equation}
\label{eq5}
    \vf_\vp^{'} = \sum_{\delta} \alpha_{\delta(\vp)}(\mW_{\delta(\vp)}^{v}) \vf_{\vp+\delta(\vp)})
    \vspace{-3pt}
\end{equation}
If we compare Eq. \ref{eq5} with Eq. \ref{eq2}, we can find the most difference between SEFormer and vanilla Transformer is that different transformations for \textit{value} points are learned based on the relative positions between points. 
% We will detail the architecture of SEFormer in next section. 

\textbf{Architecture of SEFormer}
\label{sec:seformer-arch}
Fig. \ref{fig:attention} provides a comparison between the vanilla Transformer and the proposed SEFormer. 
Given the irregularity of point cloud, we follow the paradigm of Point Transformer \cite{zhao2021point} to first sample the neighboring points around each query point independently before imported into the Transformer. 
Unlike the commonly used sampling methods like K Nearest Neighbor (KNN) or Ball Query (BQ), we choose to adopt a grid interpolation to generate the key points for Transformer. 
As shown in Fig. \ref{fig:attention}(b), around the red query point,   several (8 in Fig. \ref{fig:attention}(b)) grid arranged virtual points are generated. 
The distance between two grids is a predefined $d$. 
Then the virtual points are interpolated with their nearest neighboring points. 
Compared with traditional sampling like KNN, the advantage of grid sampling is that it can forcibly sample points from different directions.
As shown in Fig. \ref{fig:attention}(a), KNN greedily samples the nearest points. The sampled points show a strong bias from the right corner points. 
While grid interpolation can avoid such a problem and provide a better description of the local structure.
However, grid interpolation uses a fixed grid distance. 
Hence we adopt a multi-radii strategy in implementation to increase the flexibility of sampling.
% With multiple grid distance $d$, the complex point cloud structure can be well taken into account.

During calculation, in Fig. \ref{fig:attention} (a), all the anchor points share same transformation for \textit{key} and \textit{value}.
While in Fig. \ref{fig:attention} (b), SEFormer constructs a memory pool containing multiple transformation matrices ($\mW^v$) for \textit{value}. 
The interpolated key points will search their corresponding $\mW^v$ based on their relative grid coordinate to the query point. 
Then their features will be transformed differently.
For example, the top left key in Fig. \ref{fig:attention} (b) is transformed by the red $\mW^v$ while the right key is transformed by the green $\mW^v$.
In this way, SEFormer can have the structure encoding ability which is missed in vanilla Transformer.
\vspace{-5pt}
\subsection{SEFormer Based 3D Object Detection}
\label{sec:detection}
\vspace{-3pt}
The whole detection framework is shown in Fig. \ref{fig:architecture}.
We first construct a 3D convolution based backbone for multi-scale voxel features extraction and initial proposals generation.
Then a multi-scale SEFormer network is applied to extract richer local structure features for each point from the voxel features.
It contains a SEFormer based spatial structure moduling for point-level structure features and a SEFormer based ROI head for object-level structure features.

\textbf{3D CNN Backbone}
The convolution backbone first transforms the input into a series of voxel features with $1\times, 2\times, 4\times$ and $8\times$ downsampling sizes.
These differently sized features have different layers of depth.
After the feature extraction, the $8\times$ 3D feature volume will be compressed along the $Z$-axis and converted into a 2D BEV feature map.
Then a center-based approach \cite{yin2021center} is applied to predict first-stage proposals based on the BEV feature map.  
% Based on the extracted multi-level voxel features, the proposed multi-level multi-scale SEFormer structure moduling aggregates the multi-level multi-scale structure features for each point and then refines the 3D proposals.

\textbf{SEFormer Based Spatial Structure Moduling}
% \begin{figure}[h]
% \centering
% \includegraphics[width=0.47\textwidth]{figs/Fig4.pdf}
% \caption{An illustration of the proposal based query points sampling. A sphere area is set for each proposal. Points not within any sphere are filtered and query points (red points) are selected from the remaining points. }
% \label{fig:sample}
% \end{figure}
Then the proposed spatial structure moduling aggregates the multi-scale features $[\tF^{(1)}, \tF^{(2)}, \tF^{(3)}, \tF^{(4)}]$ into several point-level embedding $\tE$.
% We first use a proposal based sampling to reduce the number of embedding as in \citep{shi2021pv}. 
Starting from $\tE^{init}$, we first integrate the finest-grained features $\tF^{(1)}$. 
For each embedding point, its corresponding key points are interpolated from $\tF^{(1)}$.
We use $m$ different grid distance $d$ to generate sets of multi-scale key features as $\tF^{(1)}_1, \tF^{(1)}_2, \ldots, \tF^{(1)}_m$.
Such multi-radii strategy can better handle the sparse and irregular point distribution in LiDAR.
Then $m$ parallel SEFormer blocks which contain multiple SEFormer units are applied and result in $m$ new embedding $\tE^{(1)}_1, \tE^{(1)}_2, \ldots, \tE^{(1)}_m$.
In the end of the block, $\tE^{(1)}_1, \tE^{(1)}_2, \ldots, \tE^{(1)}_m$ are concatenated and transformed into embedding $\tE^{(1)}$ with a vanilla Transformer.
Then $\tE^{(1)}$ repeats the above process and aggregates $[\tF^{(2)}, \tF^{(3)}$, $\tF^{(4)}]$ into the final embedding $\tE^{final}$. 
Compared with the original voxel features $\tF$, the embedding $\tE^{final}$ aggregated contains richer structural description of local neighboring area.  
\begin{table*}[t]
  \centering
  \footnotesize
  \setlength{\tabcolsep}{1.8mm}{
  \scalebox{1.0}{
  \begin{tabular}{@{\extracolsep{\fill}}@{}l|c|c|ccc@{}}
    \toprule
    \multirow{2}*{Methods} & LEVEL$\_$1 (IoU=0.7) & LEVEL$\_$2 (IoU=0.7) & \multicolumn{3}{c}{ LEVEL$\_$1 3D mAP/mAPH by Distance}\\
    & 3D mAP/mAPH & 3D mAP/mAPH & 0-30m & 30-50m & 50m-Inf \\
    \midrule
    % LaserNet \cite{meyer2019lasernet} & 52.1/50.1 & -/- & 70.9/68.7 & 52.9/51.4 & 29.6/28.6 \\
    PointPillars \citep{lang2019pointpillars}* & 56.62/- & -/- & 81.01/- & 51.75/- & 27.94/- \\
    MVF \citep{zhou2020end} & 62.93/- & -/- & 86.30/- & 60.02/- & 36.02/-\\
    AFDet \citep{ge2020afdet} & 63.69/- & -/- & 87.38/- & 62.19/- & 29.27/- \\
    Pillar-OD \citep{wang2020pillar} & 69.8/- & -/- & 88.5/- & 66.5/- & 42.9/- \\
    CVCNet \citep{chen2020every} & 65.2/- & -/- & 86.80/- & 62.19/- & 29.27/- \\
    SVGA-Net \citep{he2022svga} & 73.45/- & 66.65/- & 92.53 & 69.44 & 42.08 \\
    VoTr-SSD \citep{mao2021voxel} & 68.99/68.39 & 60.22/59.69 & 88.18/87.62 & 66.73/66.05 & 42.08/41.38\\
    PV-RCNN \citep{shi2020pv} & 70.3/69.7 & 65.4/64.80 & 91.9/91.3 & 69.2/68.5 & 42.2/41.3 \\
    VoTr-TSD \citep{mao2021voxel}  & 74.95/74.25 & 65.91/65.29 & 92.28/91.73 & 73.36/72.56 & 51.09/50.01\\
    RSN \citep{sun2021rsn} $\dagger$& 75.1/74.6 & 66.0/65.5 & 91.8/91.4 & 73.5/73.1 &53.1/52.5\\
    Voxel RCNN \citep{deng2020voxel} & 75.59/- & 66.59/- & 92.49/- & 74.09/- &53.15/-\\
    SCIR-Net\citep{he2022scir} & 75.63/- & 66.73/- & 92.55/- & 72.42/- & -/- \\
    LiDAR RCNN\citep{li2021lidar}$\dagger$ & 76.0/75.5 & 68.3/67.9 & 92.1/91.6 & 74.6/74.1 & 54.5/53.4\\
    SST$\_$TS \citep{fan2022embracing} $\dagger$ & 76.22/75.79 & 68.04/67.64 & -/- & -/- & -/-\\
    CT3D \citep{sheng2021improving}  & 76.30/- & 69.04/- & 92.51/- & 75.07/- & 55.36/- \\
    Pyramid RCNN\citep{mao2021pyramid} & 76.30/75.68 & 67.23/66.68 &92.67/92.20 &74.91/74.21 & 54.54/53.45 \\
    CenterPoint\citep{yin2021center} $\dagger$ & 76.7/68.8 & 76.2/68.3 & -/- & -/- & -/-\\
    PDV \cite{hu2022point} & 76.85/76.33 & 69.30/68.81 & 93.13/92.71 & 75.49/74.91 & 54.75/53.90 \\
    Voxel-to-Point\citep{li2021voxel}  & 77.24/- & 69.77/- & \bf{93.23}/- & 76.21/- & 55.79/- \\
    PV-RCNN++\citep{shi2022pv}\textbf{} & 77.32/- & 68.62/- & -/- & -/- & -/-\\
    VoxSet \citep{he2022voxel} & 77.82/- & 70.21/- & 92.78/- & 77.21/- & 54.41/- \\
    \hline
    \bf{Ours } & \bf{79.02}/\bf{78.52} & \bf{70.31}/\bf{69.85} & 93.10/\bf{92.66} & \bf{78.07}/\bf{77.54} & \bf{57.60}/\bf{56.87}\\
    \bottomrule
  \end{tabular}}}
  \caption{Performance comparison on the Waymo Open Dataset with 202 validation sequences for the 3D vehicle detection. Only one frame is used for training and testing. * is re-implemented by \cite{zhou2020end}. 20$\%$ training data are used for most methods. While $\dagger$ denotes methods that use the whole 100$\%$ training dataset.}
  \label{tab:waymo-3d}
\end{table*}

 \textbf{SEFormer Based ROI Head} Based on the point-level embeddings $\tE^{final}$, the proposed head aggregates it into several object-level embedding to generate final proposals.
 To be specific, we first divide each proposal from the first-stage center based head into multiple cubic sub-regions and interpolate each sub-region with surrounding point-level embedding features.
 Due to the sparsity of point cloud, some sub-regions are often empty.
 Traditional works simply sum the features from the non-empty parts.
 However, the car side away from the LiDAR source is often sparse.
 Hence the relative positions of the empty sub-regions can provide a useful object-level structure feature for direction recognition.
 In contrast, the proposed SEFormer can utilize such information by embedding both the full and empty sub-regions.
 As shown in part III of Fig. \ref{fig:architecture}, a SEFormer block takes in both the empty and non-empty sub-regions and integrates their features into a proposal embedding $\tO\tE^{final}$.
 The stronger structure embedding ability of SEFormer can provide a better description of the object-level structure and then generates more accurate 3D proposals.

\section{Experiment}

In this work, we mainly evaluate the proposed SEFormer on Waymo. 
Because its large data scale can provide much more convincing evaluation than other benchmarks.
We will first introduce Waymo and describe the details of our implementation. 
Then we will compare with state-of-the-art works on Waymo Open and provide an ablation analysis for the proposed method.
\vspace{-8pt}

\subsection{Implementation Details}
\vspace{-3pt}
\textbf{Waymo Open Dataset.} The Waymo dataset contains 1000 LiDAR sequences in total. 
These sequences are further split into 798 training sequences (including around 158$k$ LiDAR samples) and 202 validation sequences (including around 40$k$ LiDAR samples). 
Waymo provides object annotations in the full 360$^\circ$ field. 
Its Official evaluation metrics include the standard 3D mean Average Precision (mAP) and mAP weighted by heading accuracy (mAPH). 
In this work, we present such two metrics mainly from two aspects: difficulty levels and object distance. For the first way, the ground-truth boxes are divided into two groups: LEVEL$\_$1 (number of points is more than five) and LEVEL$\_$2 (the box contains at least one point). 
For the second way, apart from the overall mAP, we will also show respective mAP for objects located in $0 - 30 m$,  $30 - 50 m$, and $> 50 m$.

\textbf{Network Architecture.} First, the points within the range of $[-75.2$, $75.2]m$, $[-75.2, 75.2]m$, and $[-2, 4]m$ for the X, Y, and Z-axis are extracted.
Then they are voxelized with a $(0.05m, 0.05m, 0.1m)$ step.
Our first-stage convolution backbone and the BEV neck follow the same architecture in \citep{yan2018second}. 
The 3D backbone transforms the input into $1\times, 2\times, 4\times$ and $8\times$ downsampled voxel volumes with $16,32,64,64$ dimensions respectively. 
In the SEFormer based spatial structure moduling, 4096 query points are selected for each scene. 
In the SEFormer head, each proposal is divided into  $6\times 6 \times 6$ sub-regions. 

\textbf{Training and Inference.}
We use 4 RTX 3090 GPUs to train the entire network with batch size 8.
We keep most training and inference hyper-parameters same with existing works\cite{mao2021pyramid, shi2022pv, deng2020voxel, mao2021voxel, sheng2021improving} for a fair comparison.
We adopt AdamW optimizer and one-cycle policy\cite{smith2019super} with division factor 10 and momentum ranges from 0.95 to 0.85to train the model. 
The learning rate is initialized with 0.003. 
The training time is 40 epochs. 
Given the large scale of Waymo dataset, we uniformly use $20\%$ training samples for training but use the whole validation set for evaluation. 
\vspace{-8pt}

\subsection{Detection Results on Waymo Detection Dataset}

\begin{table}[t]
  \centering
  \footnotesize
  \setlength{\tabcolsep}{0.95mm}{
  \scalebox{1.0}{
  \begin{tabular}{@{}c|c|cccc}
    \hline
    Diff. & Methods & Overall & 0-30m & 30-50m & 50m-Inf \\
    \hline
    % \multirow{11}*{LV$\_$1}&LaserNet (\citeyear{meyer2019lasernet}) & 71.57 & 92.94 & 74.92 & 48.87\\
    \multirow{11}*{LV$\_$1}&PointPillars(\citeyear{lang2019pointpillars})* & 75.57 & 92.10 & 74.06 & 55.47\\
    &MVF(\citeyear{zhou2020end}) &80.40 & 93.59 & 79.21 & 63.09\\
    &Pillar-OD (\citeyear{wang2020pillar}) & 87.11 & 95.78 & 84.87 & 72.12\\
    &PV-RCNN (\citeyear{shi2020pv}) & 82.96 & 97.35 & 82.99 & 64.97\\
    & SVGA-Net (\citeyear{he2022svga}) & 83.52 & 97.60 & 83.14 & 64.52 \\
    % &\citep{li2021anchor} & 87.15 & 97.42 & 86.80 & 73.92 \\
    &Voxel RCNN (\citeyear{deng2020voxel}) & 88.19 & 97.62 & 87.34 & 77.70\\
    &SCIR-Net (\citeyear{he2022scir}) & 88.45 & 97.71 & 88.41 & - \\
    % &RSN (\citeyear{sun2021rsn}) & 88.5 & - & - & -  \\
    &LiDAR RCNN (\citeyear{li2021lidar}) & 90.1 & 97.0 & 89.5 & 78.9 \\
    &Voxel-to-Point(\citeyear{li2021voxel}) & 88.93 & 98.05 & 88.25 & 79.19 \\
    &VoxSet(\citeyear{he2022voxel}) & 90.31 & 96.11 & 88.12 & 77.98\\
    & \bf{Ours} & \bf{91.73} & \bf{98.13} & \bf{91.23} & \bf{82.12}\\
    \hline
    \multirow{8}*{LV$\_$2} & PV-RCNN \citeyear{shi2020pv} & 77.45 & 94.64 & 80.39& 55.39 \\
    % &\citeyear{li2021anchor} & 77.85 & 94.39 & 80.87 & 59.16 \\
    & SVGA-Net (\citeyear{he2022svga}) & 80.97 & 95.54 & 81.58 & 60.18\\
    & Voxel RCNN (\citeyear{deng2020voxel}) & 81.07 & 96.99 & 81.37 & 63.26\\
    % & RSN (\citeyear{sun2021rsn}) & 81.2 & - & - & -\\
    &SCIR-Net (\citeyear{he2022scir}) & 81.65 & 96.88 & 81.34 & - \\
    & LiDAR RCNN (\citeyear{li2021lidar}) & 81.7 & 94.3 & 82.3 & 65.8\\
    &Voxel-to-Point(\citeyear{li2021voxel}) & 82.18 & 97.48 & 82.51 & 64.86 \\
    &VoxSet(\citeyear{he2022voxel}) & 80.56 & 96.79 & 80.44 & 62.37\\
    & \bf{Ours} & \bf{85.18} & \bf{97.55} & \bf{85.99} & \bf{69.48}\\
     \hline
  \end{tabular}}}
  \caption{Comparison of BEV vehicle detection on the WOD with 202 validation sequences. * is re-implemented by \cite{zhou2020end}. We only use 20$\%$ training data.}
  \label{tab:waymo-bev}
\end{table}

\begin{figure*}[t]
\centering
\includegraphics[width=1.0\textwidth]{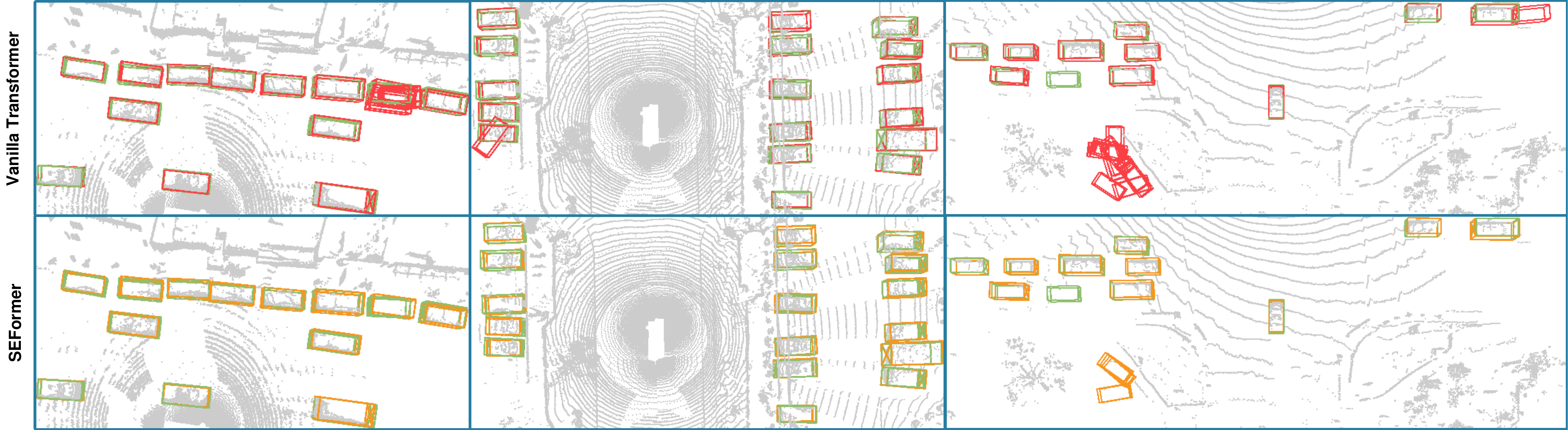}
\caption{Qualitative visualization on WOD. The green boxes denote the groundtruth.}
\label{fig:visual}
\end{figure*}

Fig. \ref{fig:visual} illustrates a qualitative presentation of our detection results on Waymo dataset.
Table \ref{tab:waymo-3d} and Table \ref{tab:waymo-bev} show the 3D and BEV vehicle precision comparison with SOTA works on the Waymo Open Dataset.
0.7 IoU threshold is adopted for both evaluations. 
We mainly compare the one-frame detection results here. 

In Table \ref{tab:waymo-3d}, it can be found further improvement of current 3D object detection has become more and more difficult.
% For example, \citet{li2021lidar} (CVPR2021) have achieved $76.0\%$LEVEL$\_$1  mAP but one newest work \cite{fan2022embracing} (CVPR2022) only improves that to $76.22\%$.
However, our SEFormer can still achieve a significant improvement compared with prior works. 
For the commonly used LEVEL$\_$1 mAP/mAPH, we achieve $79.02\%$/$78.52\%$, which exceeds state-of-the-art works by $1.2\%$ for the LEVEL$\_$1 mAP. 
For the difficult LEVEL$\_$2 result, we can still get the SOTA results and  achieve $70.31\%$/$69.85\%$ for mAP/mAPH. 
Such results demonstrate the effectiveness of the proposed SEFormer. 
To evaluate the influence of object distance, we also provide the range-based LEVEL$\_$1 mAP/mAPH.
% Given that the best existing work \cite{shi2021pv} does not provide their range based precision, we mainly compare with Voxel-to-Point \cite{li2021voxel}.
% For the overall detection precision, we behave $1.2\%$ mAP improvement ($78.85\%$ mAP vs $77.24\%$ mAP).
Although we show lower precision on near objects ($<30m$), $93.10\%$ mAP vs $93.23\%$ mAP, our improvement for distant objects is much more significant.
The improvement for $30-50m$ and $50m-Inf$ targets achieves $0.86\%$ and $3.19\%$ mAP respectively.
In most cases, the distant objects are often sparse and only show part of the outline of the objects, which makes extracting useful structure information much more difficult.
While SEFormer's  structure \textbf{preserving} and \textbf{encoding} ability alleviates such problem.

In Table \ref{tab:waymo-bev}, SEFormer also outperforms prior works on BEV precision. 
$91.73\%$ LEVEL$\_$1 and $85.18\%$ LEVEL$\_$2 BEV mAP are achieved.
It can be found that the LEVEL$\_$2 improvement is higher.
Compared with LEVEL$\_$1, LEVEL$\_$2 contains objects that have very few points.
Hence such BEV results also support the above claim that SEFormer has more advantages for sparse objects.
\vspace{-8pt}

\subsection{Ablation Study}

\begin{table}[t]
  \centering
  \footnotesize
  \setlength{\tabcolsep}{3.6mm}{
  \begin{tabular}{cc|c|c}
    \hline
    \multirow{2}*{SSM} &\multirow{2}*{Head} & LV$\_$1 (IoU=0.7) & LV$\_$2 (IoU=0.7) \\
    & & 3D mAP/mAPH & 3D mAP/mAPH\\
     \hline
    T &T & 76.10/75.61 & 68.24/67.78\\
   S &T & 77.54/77.05 & 68.82/68.38 \\
   S &S & \bf{79.02}/\bf{78.52} & \bf{70.31}/\bf{69.85}\\
   
     \hline
  \end{tabular}}
  \caption{Comparison between vanilla Transformer and SEFormer. Here \textit{SSM} and \textit{Head} respectively denote the spatial structure moduling and the detection head while \textit{T} and \textit{S} represent vanilla Transformer and SEFormer respectively.}
  \label{tab:comp}
\end{table}

\begin{table}[t]
  \centering
  \footnotesize
  \setlength{\tabcolsep}{3.6mm}{
  \begin{tabular}{c|c|c}
    \hline
    \multirow{2}*{Block Num (m)} & LV$\_$1 (IoU=0.7) & LV$\_$2 (IoU=0.7) \\
     & 3D mAP/mAPH & 3D mAP/mAPH\\
     \hline
    1 & 78.76/78.25 & 70.08/69.62\\
   2 & \bf{79.02}/\bf{78.52} & \bf{70.31}/\bf{69.85} \\
   3 & 78.81/78.32 & 70.00/69.55\\
   
     \hline
  \end{tabular}}
  \caption{Effects of the number of parallel SEFormer blocks.}
  \label{tab:groupnum}
\end{table}

\begin{table}[t]
  \centering
  \footnotesize
  \setlength{\tabcolsep}{3.8mm}{
  \begin{tabular}{c|c|c}
    \hline
    \multirow{2}*{Head Num (h)}  & LV$\_$1 (IoU=0.7) & LV$\_$2 (IoU=0.7) \\
     & 3D mAP/mAPH & 3D mAP/mAPH\\
     \hline
    1 & 78.87/78.37 & 70.14/69.69\\
   2 & \bf{79.02}/\bf{78.52} & \bf{70.31}/\bf{69.85} \\
   4 & 79.00/78.51 & 70.30/69.85\\
   
     \hline
  \end{tabular}}
  \caption{Effects of the head number in SEFormer.}
  \label{tab:headnum}
\end{table}

\begin{table}[t]
  \centering
  \footnotesize
  \setlength{\tabcolsep}{1mm}{
  \begin{tabular}{cccc|c|c}
    \hline
    \multirow{2}{*}{conv1} & \multirow{2}{*}{conv2} & \multirow{2}{*}{conv3} & \multirow{2}{*}{conv4} & LV$\_$1 (IoU=0.7) & LV$\_$2 (IoU=0.7) \\
     &&&& 3D mAP/mAPH & 3D mAP/mAPH\\
     \hline
    \checkmark & & & & 78.54/78.04 & 69.94/69.49\\
    \checkmark & & \checkmark & & 78.81/78.32 & 70.16/69.71 \\
    \checkmark& &\checkmark &\checkmark & \bf{79.02}/\bf{78.52} & \bf{70.31}/\bf{69.85} \\
    \checkmark&\checkmark & \checkmark& \checkmark& 78.77/78.29 & 69.96/69.5 2  \\
   
     \hline
  \end{tabular}}
  \caption{Effects of multi-scale features. }
  \label{tab:multi-scale}
\end{table}

\begin{table}[t]
  \centering
  \footnotesize
  \setlength{\tabcolsep}{5mm}{
  \begin{tabular}{c|c|c}
    \hline
     & LV$\_$1 (IoU=0.7) & LV$\_$2 (IoU=0.7) \\
     & 3D mAP/mAPH & 3D mAP/mAPH\\
     \hline
   w/o GI & 78.63/78.14 & 69.83/68.38 \\
   w/ GI & \bf{79.02}/\bf{78.52} & \bf{70.31}/\bf{69.85}  \\
   
     \hline
  \end{tabular}}
  \caption{Effects of grid interpolation. }
  \label{tab:grid}
\end{table}

\textbf{Comparison between vanilla Transformer and SEFormer} In this work, we propose a new Transformer, SEFormer, to encode the local spatial structure.
Hence we compare the performance of the vanilla Transformer and the proposed SEFormer in Table \ref{tab:comp}. 
The \textit{T} and \textit{S} represent vanilla Transformer and out SEFormer respectively.
In this work, we propose a SEFormer based \textit{spatial structure moduling} (SSM) and a SEFormer based \textit{head} to extract point- and object-level structure features. 
Hence we use Transformer based SSM and head as the baseline.  
It can be found vanilla Transformer only achieves $76.10\%/75.61\%$ and $68.24\%/67.78\%$ for LEVEL$\_$1 and LEVEL$\_$2 mAP/mAPH.
Replacing Transformer in SSM with SEFormer improves the performance to $77.54\%/77.05\%$ LEVEL$\_$1 mAP/mAPH and $68.82\%/68.38\%$ LEVEL$\_$2 mAP/mAPH.
If we further replace the Transformer in the head with SEFormer, we can further get $1.48\%$ LEVEL$\_$1 mAP and $1.49\%$ LEVEL$\_$2 mAP improvement respectively.
The results illustrate that the proposed SEFormer has a better ability to capture the structural features of local areas than Transformer.
In most Transformer based works, relative position encoding is often used as a method to introduce the relative spatial relationship.
Hence we use relative position encoding in both the Transformer based baseline and our SEFormer for a fair comparison.
Hence the improvement of SEFormer over Transformer shows that simple relative position encoding cannot fully encode the structure information.

\textbf{The effects of the number of parallel SEFormer blocks} As noted in Section \ref{sec:detection}, multiple parallel SEFormer blocks with different search radii are established. 
Hence, we investigate the effects of the number of parallel SEFormer blocks in Table \ref{tab:groupnum}. 
In implementation, the parallel SEFormer blocks have gradually doubled search radii.
While we set the initial radii as 0.4/0.8/1.2/2.4 for the multi-scale features.
According to the results, it can be found that 2 parallel SEFormer blocks achieve the best performance.
Increasing or decreasing the block number causes about $0.2\%$ LEVEL$\_$1 mAP reduction.

\textbf{The effects of head number} Multi-head Transformer often has better performance than single-head Transformer.
Hence, we provide an investigation of the effects of head number in Table \ref{tab:headnum}. 
It can be found that single-head SEFormer achieves $78.87\%/78.37\%$ and $70.14\%/69.69\%$ for LEVEL$\_$1 and LEVEL$\_$2 mAP/mAPH respectively. 
Adopting double-head SEFormer can reach $79.02\%/78.52\%$ LEVEL$\_$1 and $70.31\%/69.85\%$ LEVEL$\_$2 mAP/mAPH.
But the results reduce if we further increase the head number.
Hence we choose head number $h=2$ in this work.

\textbf{The effects of multi-scale features}
Table \ref{tab:multi-scale} demonstrates the effects of using multi-scale features. 
Only using the single-scale feature (conv1) only achieves $78.54\%$ and $69.94\%$ LEVEL 1 and LEVEL 2 mAP.
Introducing more features of different scales gradually improves the performance while using features of all 4 scales reduces the precision to some extent.
Please see our supplementary material for more results.

\textbf{The effects of grid interpolation}
Table \ref{tab:grid} illustrates the effects of grid interpolation. For the control group, grid interpolation is replaced with random sampling within a radius.
Please see our supplementary material for more results and discussions.
\begin{figure}
  \centering
  \includegraphics[width=0.46\textwidth]{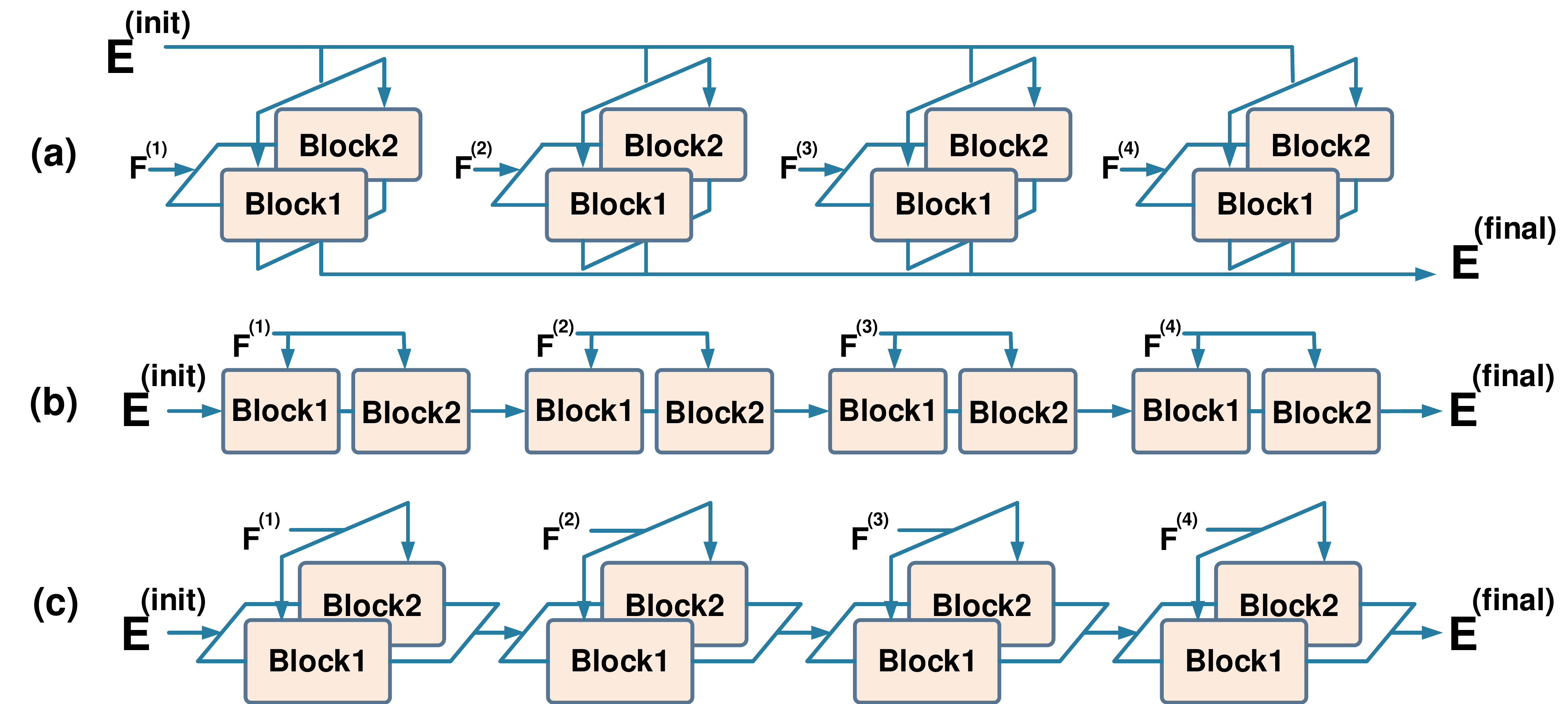}
  \caption{Illustration of (a) fully parallel (b) fully chained and (c) half parallel half chained spatial structure moduling.}
  \label{fig:aggregation}
\end{figure}

\begin{table}[t]
  \centering
  \footnotesize
  \begin{tabular}{@{}c|c|c}
    \hline
    \multirow{2}{*}{SSM Structure} & LEVEL$\_$1 (IoU=0.7) & LEVEL$\_$2 (IoU=0.7) \\
     & 3D mAP/mAPH & 3D mAP/mAPH\\
     \hline
    (a) & 78.70/78.23 & 70.04/69.56\\
    (b) & 78.86/78.39 & 70.12/69.68 \\
    (c) & \bf{79.02}/\bf{78.52} & \bf{70.31}/\bf{69.85}\\
     \hline
  \end{tabular}
  \caption{Comparison among different SSM structures.}
  \label{tab:aggregation}
\end{table}

\textbf{The structure of the spatial structure moduling}
In the spatial structure moduling, we aggregate the multi-scale features one by one. 
While multiple SEFormer blocks with different radii are adopted to extract structure information from one feature. 
To show the effects of such strategy, we design three different structures for the spatial structure moduling, fully parallel, full chained, and half parallel half chained. 
Fig. \ref{fig:aggregation} illustrates the difference among such three structures.
Half parallel half chained denotes the structure used in this work. 
Their effects on model performance are shown in Table \ref{tab:aggregation}. 
It can be found that the half parallel half chained structure has better results than others.
\section{Conclusion}
This work proposes a new Transformer, SEFormer.
In vanilla Transformer, all the value points share the same transformation.
Hence it lacks the ability to encode the distance-direction-oriented local spatial structure.
To solve such problem, SEFormer learns different transforms for value points based on their relative directions and distances to the center query point. 
Based on the proposed SEFormer, we establish a new 3D detection network including a SEFormer based spatial structure moduling to extract point-level structure information and a SEFormer based head to capture object-level structure features.
Compared with state-of-the-art solutions, our SEFormer achieves higher detection precision on the Waymo Open dataset.

% Use \bibliography{yourbibfile} instead or the References section will not appear in your paper

\bibliography{aaai23}
% \section{Acknowledgments}
% AAAI is especially grateful to Peter Patel Schneider for his work in implementing the original aaai.sty file, liberally using the ideas of other style hackers, including Barbara Beeton. We also acknowledge with thanks the work of George Ferguson for his guide to using the style and BibTeX files --- which has been incorporated into this document --- and Hans Guesgen, who provided several timely modifications, as well as the many others who have, from time to time, sent in suggestions on improvements to the AAAI style. We are especially grateful to Francisco Cruz, Marc Pujol-Gonzalez, and Mico Loretan for the improvements to the Bib\TeX{} and \LaTeX{} files made in 2020.

% The preparation of the \LaTeX{} and Bib\TeX{} files that implement these instructions was supported by Schlumberger Palo Alto Research, AT\&T Bell Laboratories, Morgan Kaufmann Publishers, The Live Oak Press, LLC, and AAAI Press. Bibliography style changes were added by Sunil Issar. \verb+\+pubnote was added by J. Scott Penberthy. George Ferguson added support for printing the AAAI copyright slug. Additional changes to aaai23.sty and aaai23.bst have been made by Francisco Cruz, Marc Pujol-Gonzalez, and Mico Loretan.

% \bigskip
% \noindent Thank you for reading these instructions carefully. We look forward to receiving your electronic files!

\clearpage
\appendix
\section{Appendix}
\subsection{More Discussion about Structure Encoding and Structure Preserving }
In this paper, we consider point cloud modeling to consist of \textbf{structure preservation} and \textbf{structure encoding}.

$\bullet$ \textbf{Structure preservation}. When capturing the structure of a local area or an object, we should only consider the related points or the points that belong to the target object. Including irrelevant points may lead to unclear object representations and thus cause inaccurate object bounding boxes.

$\bullet$ \textbf{Structure encoding} After preserving the structure of a local area or an object, we need to properly encode the structure so that the visual information is accurately embedded in representations. For point clouds, the main visual information comes from the distance and direction (or displacement) between points. Therefore, when encoding structure, the displacement information should be well reflected.

Suppose $\mP \in \mathbb{R}^{\mN \times 3}$ is the coordinates of a point cloud and $\mP \in \mathbb{R}^{\mN \times \mC}$ is the point features, where $\mN$ is the number of points and $\mC$ is the number of feature channels. Now, we elaborate on the advantage and disadvantages of standard Convolution and Transformer for point cloud structure preservation and encoding.

\textbf{\uppercase\expandafter{\romannumeral1}.Convolution is good at structure encoding but not at structure preservation.} Suppose $\mathcal{N}(\mP_i)$ denotes the neighborhood to point $\mP_i$ with a size of $l \times h \times w$. The standard Convolution models structure as follows,
\begin{equation}
    \mF_i^{'}=\sum_{\mP_j \in \mathcal{N}(\mP_i)} \mW_{ij} \cdot \mF_j
\end{equation}
where $\tW \in \mathbb{R}^{l \times w \times h \times w \times C' \times C}$ and $\mW_{ij} \in \mathbb{R}^{C' \times C}$ denotes the kernel weight for the displacement from $\mP_i$ to $\mP_j$. 
In this way, Convolution can well encode displacement because each displacement is processed by an individual kernel weight.
However, the rigid neighborhood $\mathcal{N}(\cdot)$ inevitably includes irrelevant points.
Therefore, Convolution is not good at structure preservation.

\textbf{\uppercase\expandafter{\romannumeral2}.Transformer is good at structure preservation but not at structure encoding.} The key to Transformer is self-attention, which can reflect how much two points are related.
When two points are related, the attention is high; otherwise, the attention is low. In this case, self-attention can be seen as "soft mask" which can avoid including or weaken irrelevant points.
\begin{equation}
    \mF_i^{'}=\sum_{\mP_j \in \mathcal{N}(\mP_i)} \mW \cdot \mF_j
\end{equation}
where $\mW \in \mathbb{R}^{C' \times C}$ and $\alpha_{ij}$ is the scalar attention between $\mP_i$ and $\mP_j$. However, the problem with Transformer is that it employs a shared weight $\mW$, which does not encode the displacement information.

\textbf{\uppercase\expandafter{\romannumeral3}.Our SEFormer is good at both structure preservation and structure encoding.} The proposed SETransformer combines the merits of Convolution for structure encoding and Transformer for structure preservation.
\begin{equation}
    \mF_i^{'}=\sum_{\mP_j \in \mathcal{N}(\mP_i)} \mW_{ij} \cdot \mF_j
\end{equation}
When modeling the structure centered at $\mP_i$, we treat the center point as the query point. 
Then we employ a grid interpolation to generate regular anchor points as the key points.
The self-attention is calculated based on the query point $\mP_i$ and its key points.

As mentioned before, the original Transformer is not equipped with the ability to explicitly encode the relative distance and direction (or displacement) information between the query point and its neighboring points.
In this case, Transformer fails to well sense where a neighbor is from when summing the neighbor's value.
To address this problem, we construct $3\times 3 \times 3$ grids around the query point and assign an individual mapping function to each grid.
By applying the displacement-related mapping functions to the corresponding neighbors, our SEFormer is able to know the relative positions of the neighbors, thus better capturing the spatial structure.
\subsection{Difference Between Structure Encoding and Multi-head Self-attention}
Multi-head self-attention is often adopted in current Transformer based works.
Although it also learns different mapping functions of \textit{query}, \textit{key}, and \textit{value}, we will show that the multi-head self-attention \textbf{cannot} solve the problem of structure encoding.  
The multi-head mechanism in the original Transformer can be formulated as follows,
\begin{equation}
\begin{split}
    \mF_i^{'} = [& \sum_{\mP_j \in \mathcal{N}(\mP_i)} \alpha_{ij}^{1}\mW^{1} \cdot \mF_j, \\ &\sum_{\mP_j \in \mathcal{N}(\mP_i)} \alpha_{ij}^{2}\mW^{2} \cdot \mF_j,\\
    & \ldots, \\
    & \sum_{\mP_j \in \mathcal{N}(\mP_i)} \alpha_{ij}^{h}\mW^{h} \cdot \mF_j]
\end{split}
\end{equation}
where $[\cdot]$ denotes concatenation and $h$ is the number of heads in the multi-head mechanism. 
Our SEFormer is formulated as 
\begin{equation}
    \mF_i^{'} = \sum_{\mP_j \in \mathcal{N}(\mP_i)} \alpha_{ij} \mW_{ij} \cdot \mF_j
\end{equation}
As shown in the above, although multi-head involves multiple mapping functions, they are unrelated to displacements.
Therefore, the multi-head mechanism does not well encode spatial structure.
We compare our method to the original Transformer with different numbers of heads.

As illustrated in Table \ref{tab:headnum}, using two heads achieves the best performance,
Both decreasing or increasing the head number slightly reduces the performance, about $0.2\%$ mAP difference.
Such results demonstrate that the influence of multi-head self-attention is limited. 
On the contrast, Table \ref{tab:comp} compares the effects of vanilla Transformer and the proposed SEFormer.
It can be found than replacing Transformer with SEFormer in either \textit{SSM} or the \textit{Head} achieves significant performance improvement.
There is more than $1\%$ mAP difference between SEFormer and vanilla Transformer.
Such results show the effectiveness of our SEFormer and traditional multi-head self-attention is unable to solve the structure encoding problem.
\subsection{More Results on Waymo Open Dataset}
\textbf{Detection results of other categories }. Vehicle, pedestrian and cyclist are the main three categories in outdoor 3D object detection.
Moreover, among the three categories, vehicle is the most important category.
Therefore, we mainly report the results on vehicle category in the main body.
Here, we also show the results of pedestrian detection and cyclist detection in Table \ref{tab:ped} and Table \ref{tab:cyc}, respectively.
It can be found that our SEFormer outperforms existing works in pedestrian and cyclist detection. 
\begin{table}[t]
  \centering
  \footnotesize
  \setlength{\tabcolsep}{1.2mm}{
  \begin{tabular}{c|c|c}
    \hline
    \multirow{2}{*}{Method} & LEVEL$\_$1 (IoU=0.7) & LEVEL$\_$2 (IoU=0.7) \\
     & 3D mAP/mAPH & 3D mAP/mAPH\\
     \hline
    PDV \citeyear{hu2022point} & 74.19/65.96 & 65.85/58.28 \\
    RSN \citeyear{sun2021rsn} & 77.8/72.7 & 68.3/63.7\\
    CenterPoint \citeyear{yin2021center} & 79.0/72.9 & 71.0/65.3 \\
    PV-RCNN++ \citeyear{shi2022pv} & 80.62/* & */*\\
    SEFormer (Ours) & \bf{81.08}/\bf{74.91} & \bf{72.39}/bf{66.66} \\
     \hline
  \end{tabular}}
  \caption{Pedestrian Detection on the Waymo Open Dataset}
  \label{tab:ped}
\end{table}

\begin{table}[t]
  \centering
  \footnotesize
  \setlength{\tabcolsep}{1.2mm}{
  \begin{tabular}{c|c|c}
    \hline
    \multirow{2}{*}{Method} & LEVEL$\_$1 (IoU=0.7) & LEVEL$\_$2 (IoU=0.7) \\
     & 3D mAP/mAPH & 3D mAP/mAPH\\
     \hline
    PDV \citeyear{hu2022point} & 68.71/67.55 & 66.49/65.36 \\
    PV-RCNN++ \citeyear{shi2022pv} & 73.49/* & */*\\
    SEFormer (Ours) & \bf{73.55}/\bf{72.42} & \bf{70.76}/bf{69.67} \\
     \hline
  \end{tabular}}
  \caption{Cyclist Detection on the Waymo Open Dataset}
  \label{tab:cyc}
\end{table}

\begin{table}[t]
  \centering
  \footnotesize
  \setlength{\tabcolsep}{1mm}{
  \begin{tabular}{cccc|c|c}
    \hline
    \multirow{2}{*}{conv1} & \multirow{2}{*}{conv2} & \multirow{2}{*}{conv3} & \multirow{2}{*}{conv4} & LV$\_$1 (IoU=0.7) & LV$\_$2 (IoU=0.7) \\
     &&&& 3D mAP/mAPH & 3D mAP/mAPH\\
     \hline
    \checkmark & & & & 78.54/78.04 & 69.94/69.49\\
    \checkmark & & \checkmark & & 78.81/78.32 & 70.16/69.71 \\
    \checkmark & & & \checkmark & 78.75/78.26 & 70.10/69.65\\
    & & \checkmark & \checkmark & 78.60/78.10 & 69.93/69.47 \\
    \checkmark& &\checkmark &\checkmark & \bf{79.02}/\bf{78.52} & \bf{70.31}/\bf{69.85} \\
    \checkmark&\checkmark & \checkmark& \checkmark& 78.77/78.29 & 69.96/69.5 2  \\
   
     \hline
  \end{tabular}}
  \caption{Effects of multi-scale features (Results Extension). }
  \label{tab:multi-scale-extend}
\end{table}

\begin{table}[t]
  \centering
  \footnotesize
  \setlength{\tabcolsep}{1.8mm}{
  \begin{tabular}{ccc|c|c}
    \hline
    \multirow{2}{*}{conv1} & \multirow{2}{*}{conv3} & \multirow{2}{*}{conv4} & LV$\_$1 (IoU=0.7) & LV$\_$2 (IoU=0.7) \\
     &&& 3D mAP/mAPH & 3D mAP/mAPH\\
     \hline
    & & & 78.63/78.14 & 69.83/69.38\\
     & \checkmark & \checkmark & 78.73/78.24 & 69.93/69.48 \\
    \checkmark & & \checkmark & 78.89/78.40 & 70.16/69.71\\
    \checkmark & \checkmark & & 78.94/78.45 & 70.20/69.75 \\
    \checkmark &\checkmark &\checkmark & \bf{79.02}/\bf{78.52} & \bf{70.31}/\bf{69.85} \\
   
     \hline
  \end{tabular}}
  \caption{Effects of grid interpolation (Results Extension). \checkmark shows grid interpolation is used for corresponding feature or the key points are sampled with Ball Query.}
  \label{tab:grid-extend}
\end{table}

\begin{table}
  \centering
  \footnotesize
  \setlength{\tabcolsep}{2mm}{
  \begin{tabular}{c|c|c}
    \hline
    \multirow{2}{*}{Layer Num} & LEVEL$\_$1 (IoU=0.7) & LEVEL$\_$2 (IoU=0.7) \\
     & 3D mAP/mAPH & 3D mAP/mAPH\\
     \hline
    1 & 78.45/77.97 & 69.67/69.23 \\
    2 & \bf{78.87}/\bf{78.37} & \bf{70.14}/\bf{69.19}\\
    3 & 78.69/78.21 & 69.95/69.51 \\
     \hline
  \end{tabular}}
  \caption{Effects of the number of SEFormer layers. Single-head SEFormer is adopted here.}
  \label{tab:layernum}
\end{table}

\textbf{More results about the effects of multi-scale features.}
In Table \ref{tab:multi-scale}, partial results about the effects of multi-scale features are provided.
While in Table \ref{tab:multi-scale-extend}, we show more results about the multi-scale features.
It can be found using multi-scale features has $0.48\%$ higher LEVEL$\_$1 mAP than only using single-scale feature.
Due to the structure encoding ability of convolution, the multi-scale features encodes structure information of surrounding areas of different sizes.
Hence it is easier for SEFormer to extract structure features than only utilizing the raw point cloud.
On the other hand, the proposed SEFormer still achieves excellent results with only \textit{conv1}.
This shows the effectiveness of SEFormer on feature extraction rather than highly dependent on the convolution backbone.
If we compare the results of \textit{conv1+conv3} and \textit{conv1+conv4} with the \textit{conv3+conv4} group, we can find the \textit{conv1} feature has more effects on the model performance.
The reason is that the fine-grained \textit{conv1} retains more structure information of the original point cloud.
Although the large-scale features have information from more distant objects, the fine-grained local details are inevitably lost.
Hence it becomes more difficult to extract the structure features without the fine-grained \textit{conv1}.
% However, using all 4 features show lower detection precision than only using \textit{conv1}, \textit{conv3}, and \textit{conv4}.
% The reason may be the inappropriate settings of the search radii in \textit{conv2}. 

\textbf{More results about the effects of grid interpolation.}
% In Table \ref{tab:grid}, we provide an initial evaluation of the effects of the grid interpolation.
Table \ref{tab:grid-extend} provides more results about the grid interpolation.
In implementation, we replace the grid interpolation in SEFormer with ball query.
To investigate the effects of grid interpolation on features with different scales, we also show results when the grid interpolation is dropped only for one specific scaled feature.
The results show that introducing grid interpolation brings $0.39\%$ LEVEL$\_$1 mAP and $0.38\%$ LEVEL$\_$2 mAP improvement.
Such results demonstrate that grid interpolation can better capture the structure of surrounding areas than commonly used sampling methods like ball query.
If we further investigate the effects of grid interpolation on differently scaled features, we can find the influence on \textit{conv1} is the most significant.
The point cloud is much more dense in the shallow features.
During sampling the key points, it is more likely for traditional sampling methods to repeat sampling from one certain direction. 
While the deep features are much sparser.
Hence even ball query has to sample from different directions and such problem can be alleviated.

\textbf{Effects of the SEFormer layers.}
In Table \ref{tab:layernum}, we evaluate the effects of SEFormer layer number.
Due to the limit of time, we show the results of single-head SEFormer here.
According to the Table, it can be found that adopting two SEFormer layers achieves the best performance.
Compared with the single-layer SEFormer, multi-layer strategy has better feature extraction ability.
However, 3-layer SEFormer has worse results than the 2-layer SEFormer.
Considering that the multi-scale features are concatenated one by one, increasing the layer number in each SEFormer block results in $3\times$ increment on the total layer number.
Hence it is more likely to overfit when high SEFormer layer number is adopted.

\subsection{More Visualization Results}
\begin{figure*}[t]
\centering
\includegraphics[width=1.0\textwidth]{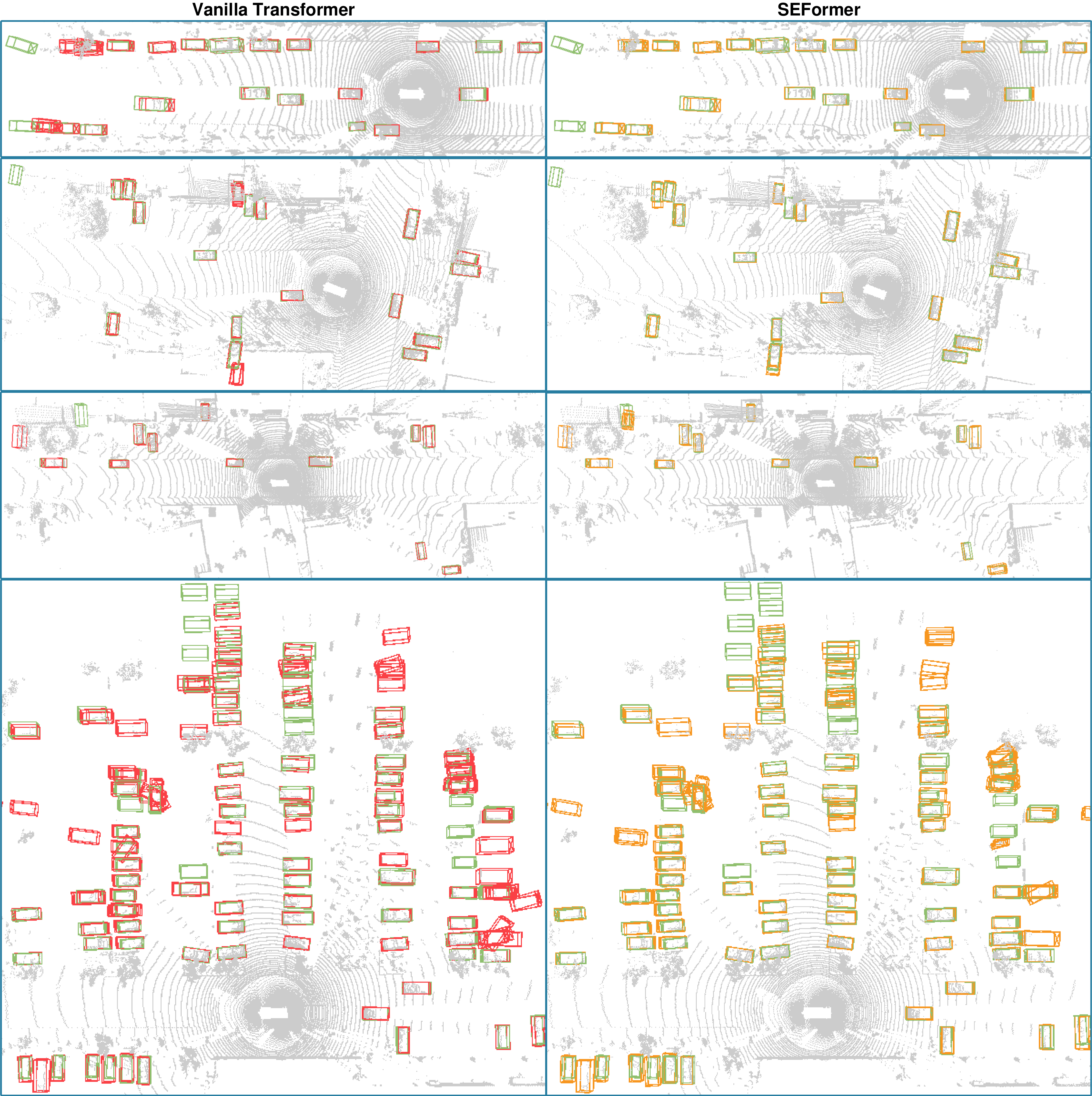}
\caption{More qualitative visualization on WOD.}
\label{fig:visual-extend}
\end{figure*}

In Fig. \ref{fig:visual-extend}, more qualitative visualization are illustrated.

\end{document}